\documentclass[review,11pt]{ReportTemplate}
\usepackage{bm}
\usepackage{graphicx}
\usepackage{paralist,algorithmic,algorithm}
\usepackage{amsmath,mathrsfs,amssymb}
\usepackage{hyperref}
\usepackage{subfigure}
\usepackage[export]{adjustbox}
\usepackage{geometry}
\begin{document}
\begin{frontmatter}
\title{Learnware: Small Models Do Big}

\author{Zhi-Hua Zhou, Zhi-Hao Tan}
\address{National Key Laboratory for Novel Software Technology\\ Nanjing University, Nanjing 210023, China\\
\rm{zhouzh@nju.edu.cn}}

\begin{abstract} 
There are complaints about current machine learning techniques such as the requirement of a huge amount of training data and proficient training skills, the difficulty of continual learning, the risk of catastrophic forgetting, the leaking of data privacy/proprietary, etc. Most research efforts have been focusing on one of those concerned issues separately, paying less attention to the fact that most issues are entangled in practice. The prevailing \textit{big model} paradigm, which has achieved impressive results in natural language processing and computer vision applications, has not yet addressed those issues, whereas becoming a serious source of carbon emissions. This article offers an overview of the \textit{learnware} paradigm, which attempts to enable users not need to build machine learning models from scratch, with the hope of reusing small models to do things even beyond their original purposes, where the key ingredient is the \textit{specification} which enables a trained model to be adequately identified to reuse according to the requirement of future users who know nothing about the model in advance.
\end{abstract} 
\end{frontmatter}

\section{Introduction}\label{sec:intro}

Machine learning has achieved great success, while there are lots of complaints about the requirement of a huge amount of training data (particularly data with labels), the difficulty of adapting a trained model to changing environments, and the embarrassment of catastrophic forgetting when refining a trained model incrementally is demanded, etc. There are great efforts such as \textit{weakly supervised learning} \cite{Zhou2018} trying to reduce the requirement of labeled training data, \textit{open-environment machine learning} \cite{Zhou2022} trying to enable learning models to adapt to environments, \textit{continual learning} \cite{Delange:Aljundi:Masana2022} trying to help deep neural networks resist forgetting; however, these issues are still far from solved.

Indeed, most efforts have been focusing on one of those concerned issues separately, paying less attention to the fact that most issues are entangled in practice. For example, a well-studied technique of weakly supervised learning for reducing the requirement of labeled training data is to collect and exploit a huge amount of unlabeled data drawn from the distribution the same as that of the labeled training data, paying less attention to the fact that in changing environments the data distributions are subject to change inherently. For another example, an effective approach to cope with changing environments is to emphasize data received in very recent timeslots since the changes have not yet caused significant differences, paying less attention to the fact that the emphasis on very recent data may tend to aggravate the severity of catastrophic forgetting.

There are many other issues, e.g., most ordinary users can hardly produce well-performed models starting from scratch, due to the lack of proficient training skills; in many real-world tasks the data privacy/proprietary issue may disable data sharing, leading to the difficulty of sharing experience among different users; in really big data applications, it is generally unaffordable or even infeasible to hold the whole data to support many passes of scanning.

The prevailing deep learning \textit{big model} paradigm, which has achieved impressive results in natural language processing and computer vision applications \cite{Vaswani:Shazeer:Parmar2017,Brown:Mann:Ryder:Subbiah2020}, has not yet addressed the above issues. Note that each big model is targeted to a task (or task class) planned in advance, generally helpless to others, e.g., a big model trained for face recognition can hardly be helpful to financial futures trading. It would be too ambitious to build a pre-trained big model for every possible task, because the number of possible tasks can be unimaginably big or even infinite. In addition, sadly, the training of big models is becoming a serious source of carbon emissions threatening our environment.

Admitting the usefulness of big models in their specifically targeted tasks, is there any paradigm offering the possibility of tackling the above issues simultaneously?

This article overviews the progress of \textit{learnware}, a paradigm offering a promising answer to the above question. It attempts to systematically reuse small models to do things that may even be beyond their original purposes, and enables users not need to build their machine learning models from scratch.

\section{The Learnware Proposal}

The learnware paradigm was proposed in~\cite{Zhou2016}. A learnware is a well-performed trained machine learning model with a \textit{specification} which enables it to be adequately identified to reuse according to the requirement of future users who know nothing about the learnware in advance.

The developer or owner \footnote{There are situations where the developer and owner of a trained machine learning model are different. Here, for simplicity, we do not distinguish them and assume that the developer holds all rights of the model.} of a trained machine learning model (no matter whether the model is a deep neural network, a support vector machine, or a decision tree, etc.) can spontaneously submit her trained model into a \textit{learnware market}. If the learnware market decides to accept the model, it assigns a specification to the model and accommodates it in the market. The learnware market should not be small, otherwise it can hardly offer help for various tasks; it would be common to accommodate thousands or millions of well-performed models submitted by different developers, on different tasks, using different data, optimizing different objectives, etc.

\begin{figure}[!t]
\begin{center}
  \includegraphics[width=.98\linewidth]{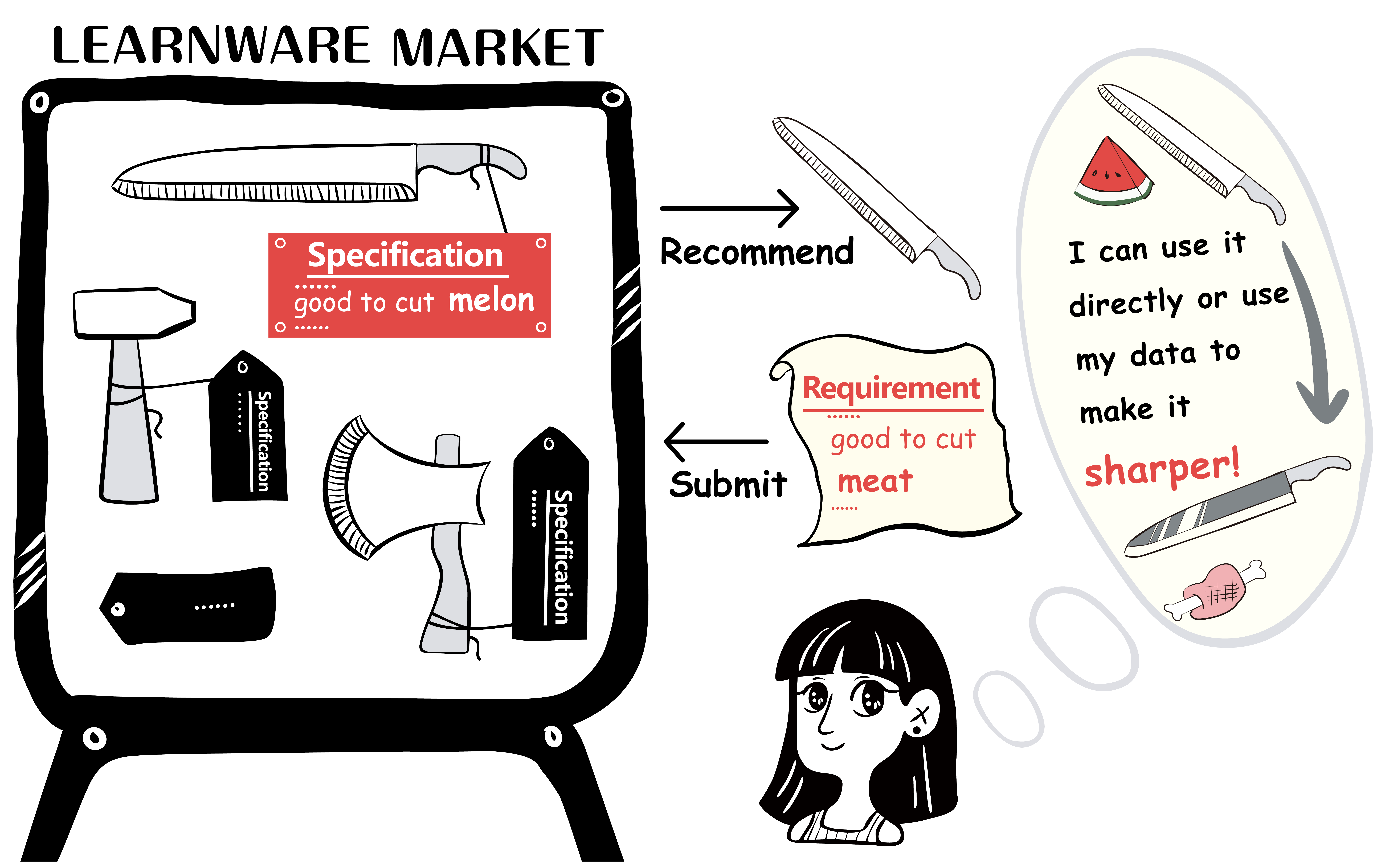}
  \caption{An analogy of learnware.}\label{fig:fig1}
\end{center}
\end{figure}

Once the learnware market has been built, when a user is going to tackle a machine learning task, she can do it in the following way rather than building her model from scratch. As the comic in Figure~\ref{fig:fig1} illustrates, she can submit her \textit{requirement} to the learnware market, and then the market will identify and deploy some helpful learnware(s) by considering the learnware specification. The learnware can be applied by the user directly, or adapted/polished by user's own data for better usage, or exploited in other ways to help improve the model built from the user's own data. No matter which mechanism for model reuse is adopted, the whole process can be much less expensive and more efficient than building a model from scratch by herself.

The learnware proposal offers the possibility of addressing most issues concerned in Section~\ref{sec:intro}:

\textbf{Lack of training data}: Strong machine learning models can be attained even for tasks with small data, because the models are built upon well-performed learnwares, and only a small amount of data are needed for adaptation or refinement for most cases.

\textbf{Lack of training skills}: Strong machine learning models can be attained even for ordinary users with little training skills, because the users can get help from well-performed learnwares rather than building a model from scratch by themselves.

\textbf{Catastrophic forgetting}: A learnware will always be accommodated in the learnware market once it is accepted, unless every aspect of its function can be replaced by other learnwares. Thus, the old knowledge in the learnware market is always held. Nothing to be forgotten.

\textbf{Continual learning}: The learnware market naturally realizes continual and lifelong learning, because with the constant submissions of well-performed learnwares trained from diverse tasks, the knowledge held in the learnware market is being continually enriched.

\textbf{Data privacy/proprietary}: The developers only submit their models without sharing their own data, and thus, the data privacy/proprietary can be well preserved. Although one could not deny the possibility of reverse engineering the models, the risk would be too small compared with many other privacy-preserving solutions.

\textbf{Unplanned tasks}: The learnware market is to be open to all legal developers. Thus, there would exist helpful learnwares in the market unless a task is new to all legal developers. Moreover, some new tasks, though no developer has built models for them specially, could be addressed by selecting and assembling some existing learners.

\textbf{Carbon emission}: Assembling small models may offer good-enough performance for most applications; thus, one may have less interest to train too many big models. The possibility of reusing other developers' models can help reduce repetitive development. Besides, a not-so-good model for one user may be very helpful for another user. No training cost wasted.

Though the learnware proposal shows a bright future, there is much work to be done to make it a reality. Section~\ref{sec:market}-\ref{sec:prototype} will present some of our progress.

\begin{figure*}[!t]
\begin{center}
  \includegraphics[width=.99\linewidth]{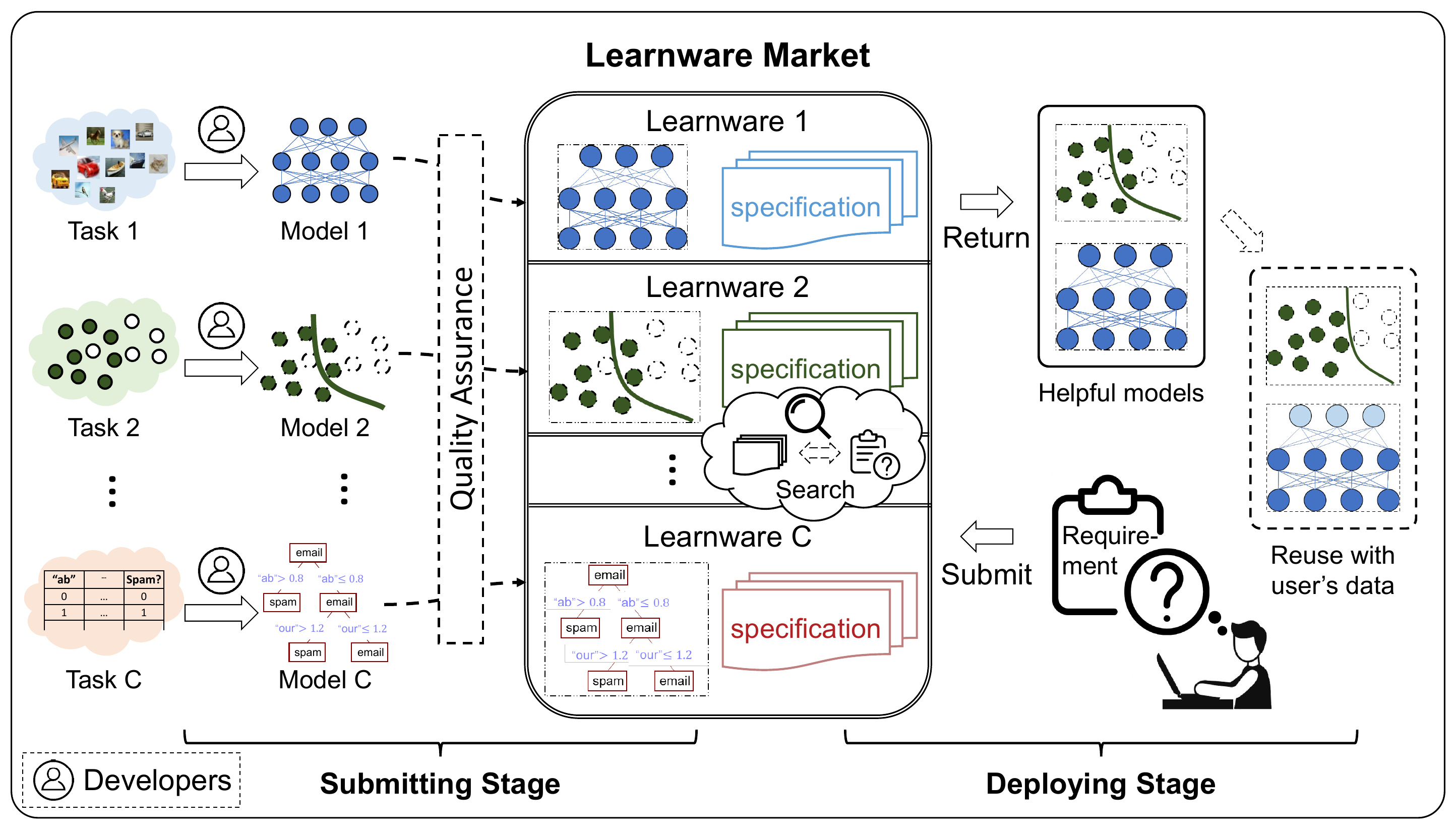}
  \caption{The learnware market and the involved two stages.}\label{fig:design}
\end{center}
\end{figure*}

\section{The Design}\label{sec:market}

There are three important entities: developers, users, and the market. The developers are usually machine learning experts who produce and want to share/sell their well-performed trained machine learning models. The users need machine learning services but usually have only limited data and lack machine learning knowledge and skills. The learnware market accepts/buys well-performed trained models from developers, accommodates them in the market, and provides/sells services to users via identifying and reusing learnwares to help users tackle their present tasks.\footnote{The learnware proposal implies some possible business relation among the three entities: The user who receives valuable services pays to the market, while market pays to developers according to the usage of their submitted learnwares. However, the business model is beyond the scope of this article.} The basic operation can be decomposed into two stages, as illustrated in Figure~\ref{fig:design}.

\subsection{Submitting Stage}\label{sec:submit}

In the submitting stage, developers can spontaneously submit their trained models to the learnware market. The market will execute some quality assurance mechanism, e.g., performance validation, to decide whether a submitted model can be accepted or not. Considering a learnware market which has accommodated millions of models, how to identify potentially helpful models for a new user?

It is evidently undesired to request the user to submit her own data to the market for trials with the models, since this would be too tedious and costly, and more seriously, this could leak user's own data. It is also impossible to utilize straightforward ideas such as ``measuring the similarity between the user data and the original training data of models'', as the learnware proposal considers the fact that neither developers nor users would like to leak their own data due to privacy/proprietary issues (it would be easier if their data are free to the market). Thus, our design is based on the constraint that the learnware market has access to neither the original training data of developers nor the original data of users. Besides, it is assumed that users know little about what models have been accommodated in the market.

The key of our solution lies in the \textit{specification}, which is the core of the learnware proposal. Once the learnware market decides to accept a submitted model, it will assign to the model a specification, which conveys the specialty and utility of the model in some format, without leaking its original training data. For simplicity, consider models corresponding to functions realizing mappings from the input domain $\mathcal{X}$, to the output domain $\mathcal{Y}$, with regard to the objective $\tt{obj}$; in other words, those models reside in a functional space $\mathcal{F}: \mathcal{X} \mapsto \mathcal{Y}~w.r.t.~ \tt{obj}$. Each model has a specification. All specifications form a \textit{specification space} where those of models that are helpful for the same tasks are nearby.

In a learnware market, there will exist heterogeneous models with different $\mathcal{X}$, and/or different $\mathcal{Y}$, and/or different $\tt{obj}$. If we call the specification space covering all possible models in all possible functional spaces as the specification \textit{world} analogically, then each specification space corresponding to one possible functional space can be called a specification \textit{island}. Designing an elegant specification format covering the whole specification world, and enabling all possible models to be efficiently and adequately identified is a grand challenge. Currently, we employ a practical design as follows. The specification of each learnware consists of two parts, where the first part explains which specification island the learnware locates, while the second part, to be introduced in Section~\ref{sec:spec}, discloses at which location it resides in this island.

\begin{figure*}[!t]
\begin{center}
  \includegraphics[width=.98\linewidth]{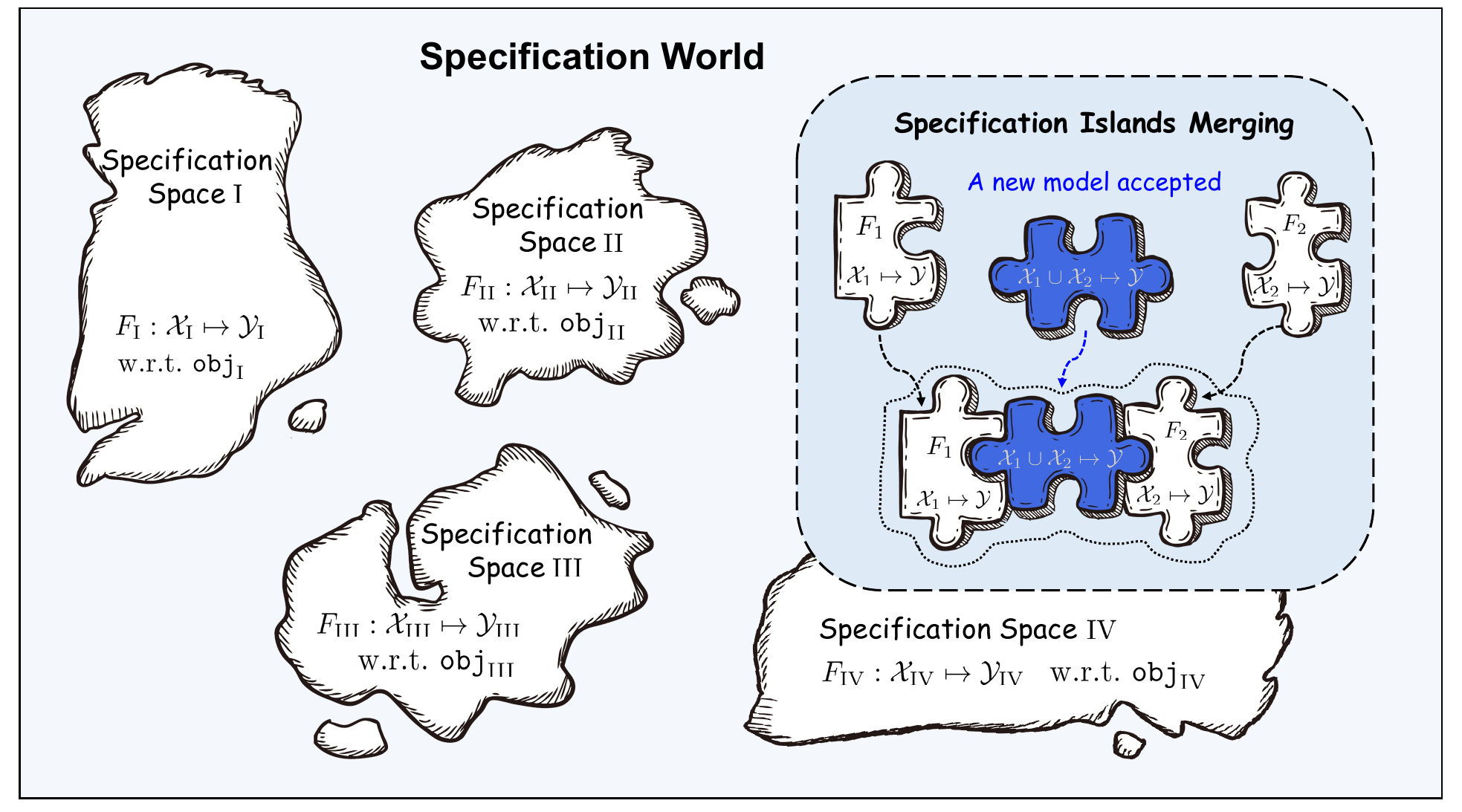}
  \caption{The learnware specification world.}\label{fig:world}
\end{center}
\end{figure*}

The first part can be realized by a string, consisting of a set of descriptions/tags given by the learnware market, about the task, input, output, and objective, etc. Then, according to the descriptions/tags provided in the user requirement, the corresponding specification island can be efficiently and accurately located. Generally, the designer of the learnware market can compose a set of initial descriptions/tags, and the set can grow when the market accepts some new models that could not be accommodated in existing functional spaces, resulting in the creation of new functional spaces and their corresponding specification islands. The learnware market can be ever-increasing as long as its host resource allows.

The specification islands can merge into a larger one, as illustrated in Figure~\ref{fig:world}. Initially there are two islands, corresponding to functional spaces $F_1: \mathcal{X}_1 \mapsto \mathcal{Y} ~w.r.t.~ \tt{obj}$ and $F_2: \mathcal{X}_2 \mapsto \mathcal{Y} ~w.r.t.~ \tt{obj}$, respectively. When a new model about $F: \mathcal{X}_1 \cup \mathcal{X}_2 \mapsto \mathcal{Y} ~w.r.t.~ \tt{obj}$ is accepted by the learnware market, these two islands can be merged. For example, suppose there are some models on text data (i.e., on $\mathcal{X}_1$) and image data (i.e., on $\mathcal{X}_2$), respectively; once some multi-modal models involving both texts and images are accepted, these text-only and image-only models can become helpful to each other, and they appear to reside in the same extended functional space (in other words, these text-only and image-only models will be kept along with the new multi-modal models). Note that though the learnware market does not have access to the original training data of models, this is still possible because the market can have synthetic data by randomly generating some inputs and feeding them to models, and then concatenating each input with its corresponding output to construct a data set reflecting the function of a model. In principle, specification islands can be merged if there are common ingredients in $\mathcal{X}$, $\mathcal{Y}$, and $\tt{obj}$. One can imagine when all possible tasks are present, all the specification islands become connected to a non-fragmented unified specification world.

\subsection{Deploying Stage}

In the deploying stage, the user submits her requirement to the learnware market, and then the market will identify and return some helpful learnwares to the user.  There are two issues, i.e., how to identify learnwares matching the user requirement, and how to reuse the returned learnwares.

The learnware market can accommodate thousands or millions of models. Different to previous machine learning studies about model reuse~\cite{Yang:Zhan:Fan:Jiang:Zhou2017,Zhao:Cai:Zhou2020} or domain adaptation~\cite{Mansour:Mohri:Rostamizadeh2009} where all pre-trained models are assumed to be helpful, there may be only a tiny portion of learnwares helpful for the current user task. Different from multi-task learning \cite{Zhang:Yang2022} where data of the multiple tasks are available in training, and domain-agnostic learning \cite{Peng:Huang:Sun2019} where labeled data of the source domain are available, learnware market does not assume to have those information. Indeed, efficiently identifying helpful learnwares is quite challenging, particularly when considering the fact that the learnware market has access to neither the original training data of learnwares nor the original data of current user.

With the specification design mentioned in Section~\ref{sec:submit}, the learnware market can request user to describe her intention using the set of descriptions/tags, through a user interface or a kind of \textit{learnware description language} to be designed in the future. Based on such information, the task reduces to how to identify some helpful learnwares in a specification island. The learnware market can provide several \textit{anchor} learnwares, such as prototypes of functionally similar learnware clusters~\cite{Xie:Tan:Jiang:Zhou2023} in the functional space corresponding to the specification island, request user to test them and return some information, and then identify potentially helpful learners based on these information, as to be explained in Section~\ref{sec:spec}. The efficiency and scalability of the learnware market can be further improved by maintaining a \textit{specification index} facilitated with relevant techniques such as hashing. Flexible specification index helps enable the learnware market to be evolvable, such that the market can accommodate an ever-increasing number of learnwares and achieve an increasingly accurate model characterization and identification, sometimes even ascertain the capabilities of models beyond their original purposes.

Once some helpful learnwares are identified and delivered to user, they can be reused in various ways. The most straightforward one is to apply the received learnware to user's own data directly; if multiple learnwares are received, they can be used to comprise an ensemble~\cite{Zhou2012} for even better performance. The user can also adapt/polish the received learnware(s) by generating a model from her own training data and putting it to use together with the learnwares(s). Another possible usage is to regard each received learnware as a feature augmentor, by feeding user's data to the learnware, taking its output for each instance as an augmented feature, and then utilizing the augmented data to build the final model.

Note that some helpful learnwares may be trained from tasks that are not exactly the same as the user's current task. For example, there are cases where learnwares with different objectives can be reused to help user's current task, such as that a model which optimizes \textit{accuracy} can be reused to help a task optimizing \textit{AUC}, by augmenting a $\delta$ function based on user's own data \cite{Li:Tsang:Zhou2013}. There are also cases where there is no single learnware that can tackle user's task as a whole, but there are multiple learnwares each can tackle a part of user's task separately. In such cases, user task can be tackled in a divide-and-conquer way, as proposed in reusable ensemble~\cite{Zhou2012}[pp.184]. Besides, if a very small set of potentially helpful learnwares have been returned, it is possible to reuse them collectively through measuring the utility of each model on each testing instance \cite{Ding:Zhou2020}. Sometimes user may find it difficult to express her requirement accurately. In such case, it is appealing to reuse the received learnware(s) by adapting/polishing them directly using user's own data. There are preliminary studies that might be somewhat helpful for this purpose, e.g.,~\cite{Li:Tsang:Zhou2013,Kuzborskij:Orabona2016,Yang:Zhan:Fan:Jiang:Zhou2017,Zhao:Cai:Zhou2020}. Moreover, in learnware deploying it would be beneficial to leverage diverse variants and hardware resource exploitation mechanisms to improve cost efficiency like in some recent explorations \cite{Romero:Li:Yadwadkar2021}.

\subsection{Learnware Specification}\label{sec:spec}

The learnware specification, ideally, should express/encode important information about every model accommodated in the learnware market to enable them to be identified efficiently and adequately for future users. As mentioned in Section~\ref{sec:submit}, our current specification design consists of two parts. The first part is a string of descriptions/tags given by the learnware market, based on information submitted by developers, aiming to locate the specification island in which a model resides. Different learnware market enterprises may employ different descriptions/tags.

The second part of the specification plays a crucial role in locating the appropriate place in the functional space $F: \mathcal{X} \mapsto \mathcal{Y} ~w.r.t.~ \tt{obj}$ for the model. Our recent effort is the RKME (Reduced Kernel Mean Embedding) specification, based on techniques of reduced set of KME (Kernel Mean Embedding) \cite{Scholkopf:Mika:Burges:Knirsch:Muller1999,Berlinet:ThomasAgnan2011}. The KME is a powerful technique to map a probability distribution to a point in RKHS (Reproducing Kernel Hilbert Space), whereas the reduced set reserves the ability with a concise representation which does not expose the original data.

Suppose a developer is to submit a model trained from data set $\{ (\bm{x}_i, \bm{y}_i)\}_{i=1}^m$, $\bm{x}_i \in \mathcal{X}$, $\bm{y}_i \in \mathcal{Y}$. Once the model is trained, the $\bm{x}_i$'s can be fed to the model to get corresponding output $\hat{\bm{y}}_i$. Note that $\hat{\bm{y}}_i$ is the output of the model instead of ground-truth, and thus, the data set $\{ (\bm{x}_i, \hat{\bm{y}}_i)\}_{i=1}^m$ encodes the function of the model; in other words, it offers a function representation of the model. Note that $\bm{x}$'s in addition to $\bm{x}_i, ~i \in [1,m]$ can also be generated and fed to the model for a more thorough representation. This idea has been explored in \cite{Zhou:Jiang2004} for learning a simpler model with comparable or even better performance from an original complicated model; a similar idea later has been called \textit{knowledge distillation} \cite{Hinton2015}. Here, we take it as the basis for constructing the first part of the RKME specification. For simplicity, let $\bm{z}_i$ denote $(\bm{x}_i, \hat{\bm{y}}_i)$, and the function of the model is encoded in the distribution of $\bm{z}_i$. Then, the market will generate the reduced set representation by minimizing the distance measured by the RKHS norm~\cite{Wu:Xu:Liu:Zhou2023}, as
\begin{equation}
  \min _{\bm{\beta}, \bm{t}}\left\|\frac{1}{m} \sum_{i=1}^{m} k\left(\bm{z}_{i}, \cdot\right)-\sum_{j=1}^{n} \beta_{j} k\left(\bm{t}_{j}, \cdot\right)\right\|_{\mathcal{H}}^{2}
\end{equation}
where $k(\cdot,\cdot)$ is the kernel function corresponding to the RKHS $\mathcal{H}$, $n \ll m$, both decided by the learnware market and given to the developer. The solved $({\bm{\beta}, \bm{t}})$, which offers a much more concise representation very different from the original data $\bm{z}$, will be submitted by the developer for the second part of the model specification.

In the deploying stage, if the user has many training data, the market can help her construct the RKME requirement to submit. Then, by matching the RKME specifications with the user requirement, the market can identify and return the learnware with the smallest distance in the RKHS norm. The market can also identify multiple helpful learnwares whose weighted combination of RKME specifications has the smallest distance to the user requirement. If the user does not have sufficient training data for constructing an RKME requirement, the learnware market can send several anchor learnwares to the user. By feeding her own data to these anchor learnwares, some information such as (precision, recall) or other performance indicators, can be generated and returned to the market. These information could help the market identify potentially helpful models, e.g., by identifying models that are far from anchors exhibiting poor performance whereas close to anchors exhibiting relatively better performance in the specification island.\footnote{Both above processes can be realized with a user interface at user's side, without leaking user's training data to the learnware market.}

Note that in the procedures described above, neither the training data of developers nor that of users are leaked to the learnware market.

\section{Some Theoretical Results}\label{sec:theo}

The RKME specification is based on RKME $\widetilde{\Phi}$, which aims to make a good representation by constructing a reduced set to approximate the empirical KME $\Phi=\int_{\mathcal{X}} k(\bm{x}, \cdot) \mathrm{d} P(\bm{x})$ of the underlying distribution. Theoretically, when the kernel function satisfies $k(\bm{x}, \bm{x}) \leq 1$ for all $\bm{x} \in \mathcal{X}$, with probability at least $1-\delta$, we have the guarantee that~\cite{zhang:yan:zhao:zhou2021,Muandet:Fukumizu:Sriperumbudur2017,Wu:Xu:Liu:Zhou2023}
\begin{equation}
  \left\|\widetilde{\Phi}-\Phi\right\|_{\mathcal{H}} \leq 2 \sqrt{\frac{2}{n}}+\sqrt{\frac{1}{m}}+\sqrt{\frac{2 \log (1 / \delta)}{m}},
\end{equation}
where $n, m$ are the size of the RKME reduced set and the original data, respectively. It is known that when using characteristic kernels such as the Gaussian kernel, KME can capture all information about the distribution~\cite{Sriperumbudur:Fukumizu:Lanckriet2011}. Besides, when the RKHS of the kernel function is finite-dimensional, RKME enjoys a linear convergence rate $O\left(e^{-n}\right)$ to empirical KME~\cite{Bach:Lacoste-Julien:Obozinski2012}; even for infinite-dimensional RKHS, it has been proved constructively that RKME can enjoy $O\left(\sqrt{d}/n\right)$ convergence rate under $L_{\infty}$ measure, where $d$ is the dimension of original data~\cite{Karnin:Liberty2019,Phillips:Tai2020}. Therefore, the RKME is guaranteed to be a good estimation of KME and a valid representation for data distribution that encodes the ability of a trained model.

The risk on the user task can be bounded under some assumptions, such as the assumption that the distribution corresponding to the task of user matches that of a learnware, or the assumption that it can be approximated by a mixture of distributions corresponding to a set of learnwares' tasks, i.e.,
\begin{equation}
\mathcal{D}_{u}=\sum_{i=1}^{N} w_{i} \mathcal{D}_{i} \ ,
\end{equation}
where $\mathcal{D}_u$ is the distribution corresponding to user task, $N$ is the number of learnwares and $\mathcal{D}_i$ are their corresponding distributions, $\sum_{i=1}^{N} w_{i}=1$ and $w_{i} \geq 0$. These two assumptions are called \emph{task-recurrent} and \emph{instance-recurrent} assumptions respectively \cite{Wu:Xu:Liu:Zhou2023}. Besides, assume that all learnwares are well-performed ones, i.e.,
\begin{equation}
  \mathbb{E}_{ \mathcal{D}_i}\left[\ell(\widehat{f_{i}}(\bm{x}), \bm{y})\right] \leq \epsilon, \ \forall i\in[N],
\end{equation}
where $\widehat{f_{i}}$ is the function corresponds to the $i$-th learnware, $\ell$ is the loss function, $\bm{y}$ is assumed to be determined by a ground-truth global function $h$.

Under these assumptions, recent studies have attempted to bound the risk on user task \cite{Wu:Xu:Liu:Zhou2023,zhang:yan:zhao:zhou2021}. Consider the task-recurrent assumption and select the learnware $\left(\widehat{f}_{i}, \widetilde{\Phi}_{i}\right)$ with the smallest RKHS distance $\eta$ according to RKME, given the loss function
\begin{equation}
\left| \ell(\widehat{f_{i}}(\bm{x}), h(\bm{x})) \right| \leq U , ~\forall \bm{x}\in\mathcal{X}, ~\forall i\in[N],
\end{equation}
we have the following result for empirical risk on user task
\begin{equation}
  \widehat{\mathbb{E}}_{\mathcal{D}_u}\left[\ell(\widehat{f_{i}}(\bm{x}), \bm{y})\right] \leq \epsilon+U \eta + O\left(\frac{1}{\sqrt{m}}+\frac{1}{\sqrt{n}}\right) \ .
\end{equation}
%\begin{equation}
%  \mathbb{E}_{\mathcal{D}_u}\left[\ell(\widehat{f_{i}}(\bm{x}), \bm{y})\right] \leq \epsilon+U \eta + O(\frac{1}{\sqrt{m}}+\frac{1}{\sqrt{m_{t}}}+\frac{1}{\sqrt{n}}),
%\end{equation}
%where $m_t$ is the data size of user task and $n$ is the size of all RKME reduced sets.
As for the instance-recurrent assumption and the $0/1$-loss
\begin{equation}
\ell_{01}(f(\bm{x}), \bm{y})=\mathbb{I}(f(\bm{x}) \neq \bm{y}),
\end{equation}
a more general result of generalization error on user task has been attained~\cite{zhang:yan:zhao:zhou2021}:
\begin{equation}
  \mathbb{E}_{\mathcal{D}_u}\left[\ell_{01}(f(\bm{x}), \bm{y})\right] \leq \epsilon+R(g),
\end{equation}
where $R(g)=\sum_{i=1}^{N} w_{i} \mathbb{E}_{\mathcal{D}_{i}}\left[\ell_{01}(g(\bm{x}), i)\right]$ represents the weighted risk of any learnware selector $g(\bm{x})$, which takes unlabeled data as input and assigns it to the proper model, $f(\bm{x})=\widehat{f}_{g(\bm{x})}(\bm{x})$ is the final model for user task.
%This indicates that the generalization risk on user task could be bounded by the model performance on original tasks, and heavily determined by the quality of learnware reuse selector.

There are efforts trying to enable the learnware market to handle unseen jobs \cite{zhang:yan:zhao:zhou2021}, where the user task involves some unseen parts that have never been handled by current learnwares in the market, and a more general theoretical analysis is presented based on mixture proportion estimation \cite{Ramaswamy:Scott:Tewari2016,Plessis:Niu:Sugiyama2017}.

There are studies in the field of ensemble learning \cite{Zhou2012} showing that assembling a group of weak machine learning models can reach strong performance; e.g., it has been proven that \textit{weak learnability} equals to \textit{strong learnability} \cite{Schapire1990}, implying that small weak models can be boosted into a strong model if adequately assembled. For a learnware market, however, it is not wise to assume that for arbitrary specific applications there must exist in this market some models that can be exploited to offer strong performance. In practice, the anchor learnwares can offer an indicator, e.g., if the performance of all anchor learnwares are very poor then it may suggest that there is little hope to find helpful learnwares in this market, while theoretical study remains to be explored.

\section{A Simple Prototype}\label{sec:prototype}

\begin{figure*}[!t]
\begin{center}
  \includegraphics[width=.95\linewidth, frame]{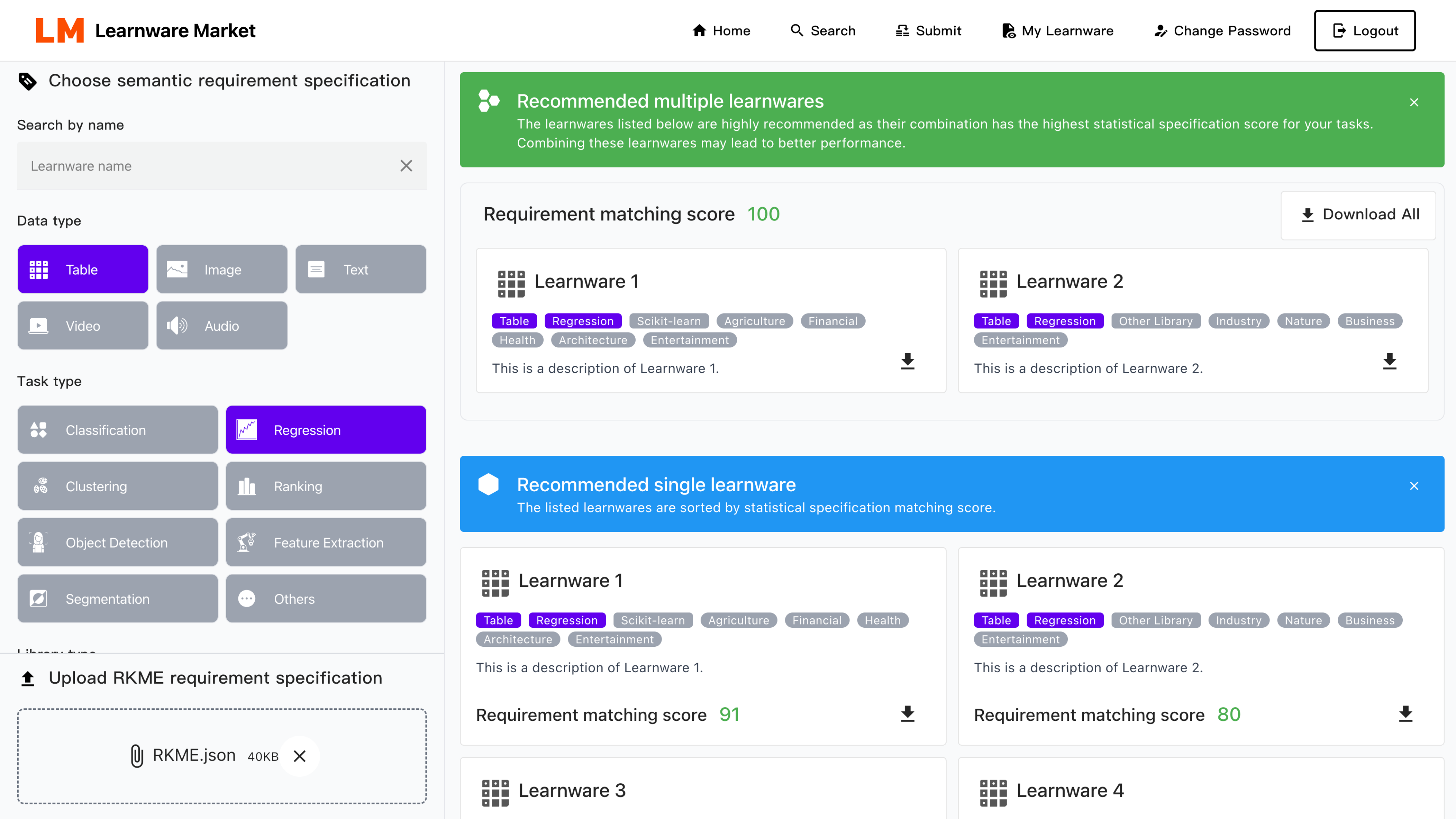}
  \caption{Interface of the simple prototype.}\label{fig:interface}
\end{center}
\end{figure*}

A simple prototype learnware market has been implemented for experiments,\footnote{An open-source experimental system with documentation will be made available soon.} with an interface shown in Figure~\ref{fig:interface}: the left-hand panel is for user to submit requirement specification (including semantic part and/or RKME part), while the right-hand panel returns learnwares identified from the market, with \textit{requirement matching scores} showing how well the learnwares match the user requirement estimated via RKME specification.

\begin{figure}[!t]
\begin{center}
  \includegraphics[width=.65\linewidth]{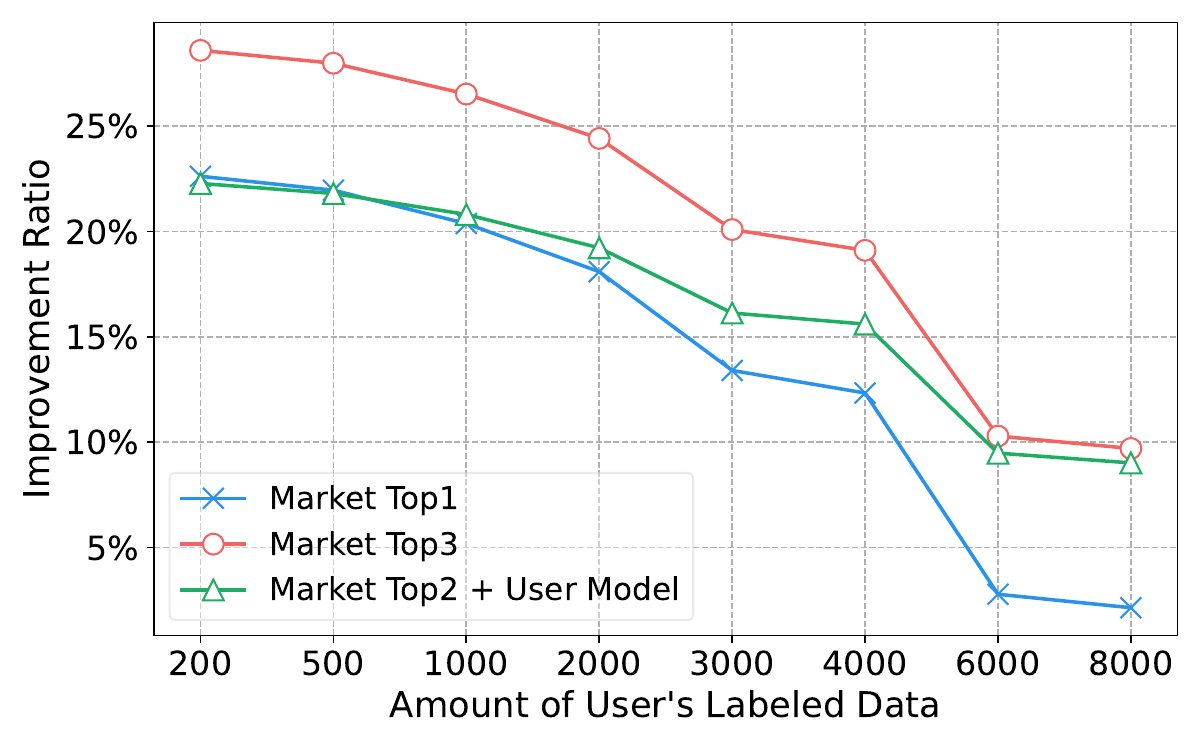}
  \caption{Sales forecasting task: Benefit from learnware market in contrast to building a model from scratch using user's own data.}\label{fig:sales}
\end{center}
\end{figure}

The market accommodates 53 models about sales forecasting. They are with different model types and trained from different data sets, though the input space, output space and objectives are the same. Thus, the specifications of these models reside in the same specification island realized with the RKME specification. Experiments are conducted to simulate the scenario where a new user, who plans to build her own sales forecasting model, is to get help from the learnware market.
The following approaches are tried: to use the best-single model identified from the market directly, to use the best-three models returned from the market via ensemble averaging, to use the best-two models returned from the market together with a model trained from user's own data via ensemble averaging, respectively. The models returned from the learnware market are identified based on RKME specification matching. Figure~\ref{fig:sales} shows the performance improvement ratio of those approaches against that of the model trained by using user's own data only. Note that the user model was developed with rich machine learning expertise, and when the user has 8,000 labeled data, the user model performance is highly competitive with the best model in learnware market.

Figure~\ref{fig:sales} exhibits that by resorting to the learnware market, the user can get much better models than simply building a model from scratch by using her own data, especially when she has only a small amount of labeled data. In particular, when the user has only 200-1,000 labeled data, using the best-single model identified from the learnware market (i.e., Top1) brings more than 20\% performance improvement. This verifies that the learnware paradigm can offer a remedy to the lack of training data. It is also noteworthy that when a group of models is identified from the learnware market (i.e., Top3), the improvement against the user's own model is always apparent. Even when the user has 8,000 labeled data such that the performance of her own model is highly competitive to the best model and better than the second-best model in the market, these models from learnware market can still be helpful as 10\% improvement is observed when they are used together with user's own model (i.e., Top2+User model).

\begin{figure}[!t]
\begin{center}
  \includegraphics[width=.65\linewidth]{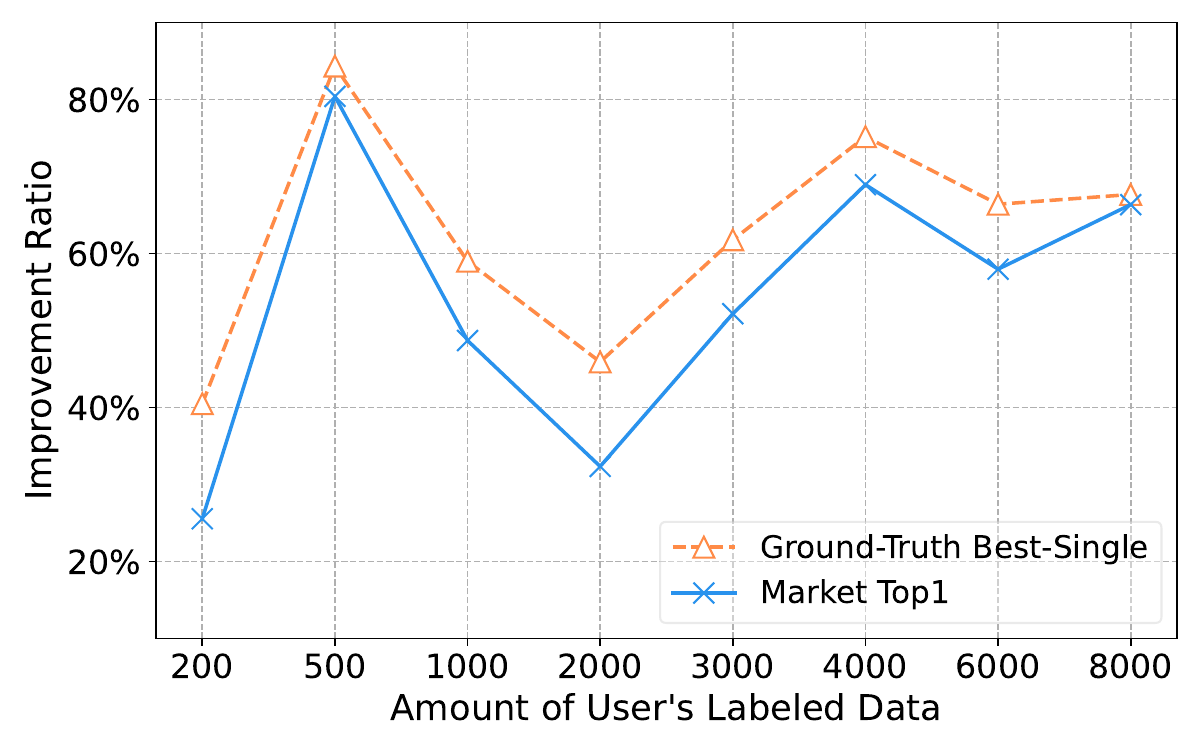}
  \caption{Performance difference of the best-single model (ground-truth) and the Top1 model (identified according to RKME specification matching) in contrast to average model (five random runs).}\label{fig:matching}
\end{center}
\end{figure}

Figure~\ref{fig:matching} plots the performance of the Top1 model identified according to RKME specification matching in the above experiments, the ground-truth best-single model in the learnware market, against the average model performance from five random runs of selection. It can be seen that the performance of the Top1 model is far superior to that of average model, and quite close to the ground-truth best-single model in most cases, verifying that the RKME specification matching is effective.

\begin{figure}[!t]
\begin{center}
  \includegraphics[width=.65\linewidth]{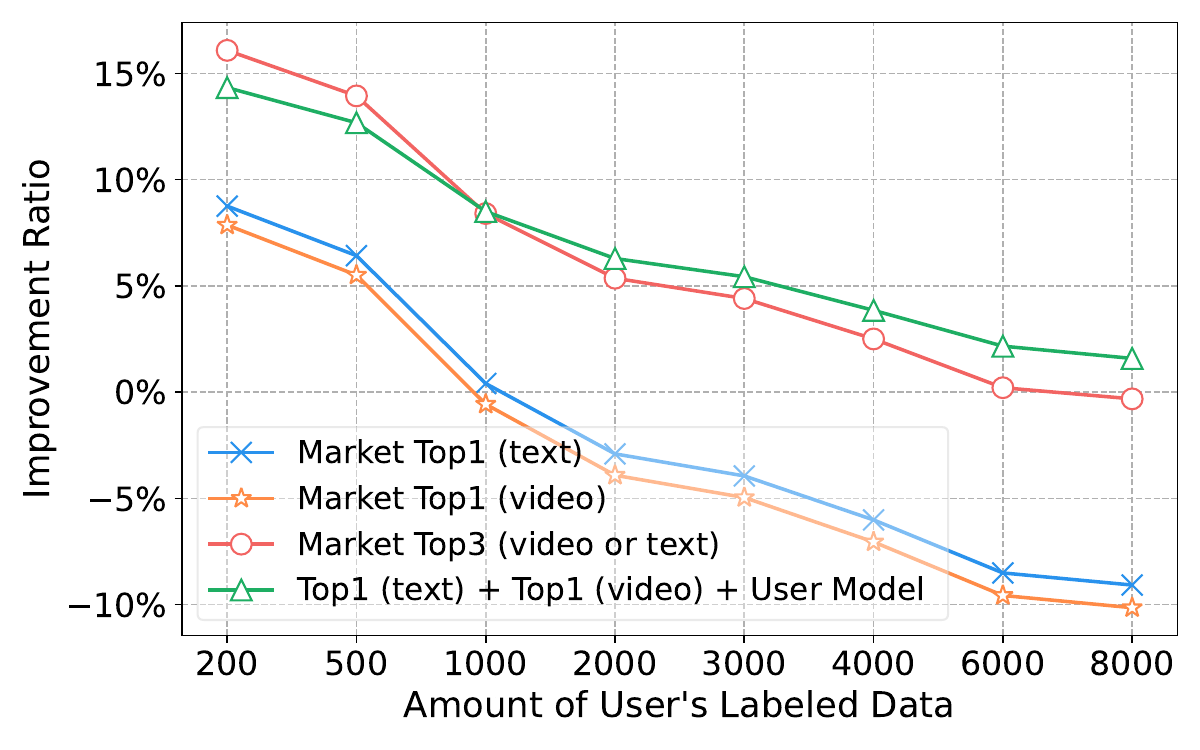}
  %\captionsetup{justification=centering}
  \caption{Sentiment analysis task: Benefit from learnware market in contrast to building a model from scratch using user's own data.}\label{fig:sentiment}
\end{center}
\end{figure}

The prototype learnware market also accommodates 6 models about sentiment analysis on video data, and 6 models on textual data with the same output space. A new user, who has some data involving both video and textual information, resorts to the learnware market, and the following approaches are tried: to use the best-single video-only or text-only model identified from the market, to use these two models together with a model trained from user's own data via ensemble averaging, to use the best-three models (no matter whether they are video-only or text-only) returned from the market via ensemble averaging, respectively. Figure~\ref{fig:sentiment} shows the performance improvement ratio of those approaches against that of the model trained by using user's own data only.

Figure~\ref{fig:sentiment} exhibits again that by resorting to the learnware market, the user can get much better results than using her own data only to train a model, especially when she has just a small amount of data. In particular, it can be seen from the figure that when user has more than 1,000 labeled data, the improvement of the best-single models identified from learnware market (i.e., Top1(text) or Top1 (video)) against the user model are both negative, showing that user's own model is better than the best-single models identified from the learnware market; this is not strange because this is a multi-modal task whereas none of models in the current learnware market was developed for multi-modal tasks. However, it is amazing that when user has less than 1,000 labeled data, either the Top1 (text) or Top1 (video) model can do 5\% or even 10\% better than user model despite the fact that they were not developed for multi-modal tasks whereas the user model was trained exactly for the current multi-modal task. If the three best-single models identified from learnware market (i.e., Top3) are used, though just by the simple ensemble averaging mechanism, performance improvement is visible even when user has 5,000 labeled data, despite the fact that none of the Top3 models were trained for multi-modal tasks. These observations verify our argument that learnwares can be useful beyond their original purposes. Furthermore, performance improvement is always visible if user model is employed together with the Top1 video-only and text-only models, implying that it is beneficial to exploit user's own data to adapt/polish the models obtained from learnware market.

It is worth highlighting that such a task, i.e., building a model for sentiment analysis from multi-modal data involving both video and textual information, has never been tackled by previous developers and no model for the exact task exists in the learnware market. This verifies that some new tasks, though no developer has built model for them specifically, can be addressed by selecting and assembling some existing learners.
Note that once the user model on multi-modal data is submitted to the learnware market, the two specification islands that correspond to video-only and text-only models, respectively, will merge just like that illustrated in Figure~\ref{fig:world}.

Though our simple prototype and experiments have exhibited promising aspects of the learnware paradigm, they are still preliminary and thorough empirical study is left for future when large-scale especially enterprise-level learnware markets are available.

\section{Conclusion and Future Issues}

This article provides a brief overview of progress on \textit{learnware}, a paradigm that seems promising to tackle many concerns of current machine learning techniques, such as the lack of training data and skills, catastrophic forgetting, continual learning, data privacy/proprietary preserving, unplanned tasks, carbon emission, etc. It would be great if, in the future, users who plan to build their own machine learning models would look into the learnware market first rather than starting from scratch themselves, just like today's programmers looking for useful codes from Github or other codebases.

There are too many issues for future exploration. First, ideally, the learnware specification should enable well-performed models helpful for the same tasks to locate nearby, whereas our current design is making models with similar functions locate nearby. Considering a user task which can be collectively tackled by several learnwares, one possibility is to tackle the task in a divide-and-conquer way and then look for helpful learnwares for each sub-task. This is to be explored in future. Second, currently the learnwares are assumed to be based on well-performed models whose function can be represented by its training data distribution, whereas in practice the models submitted by developers can be less well-performed. The quality assurance as well as its influence on the identification and reuse procedures are to be studied in future. Third, when the user does not have sufficient data for distribution estimation, as mentioned in Section~\ref{sec:spec}, some anchor learnwares are to be sent to user. This can be realized by selecting prototype models through functional space clustering, and more interesting designs are to be explored in future. Note that our current design tries to assign each model to one location. It is, however, often the case that one model can be helpful for a variety of tasks. To enable one model to be located in multiple suitable specification islands simultaneously is another interesting future issue. Besides, to explore various ways to merge specification islands is also interesting. The learnware market also offers a platform to study the possible ``intellectual ability emergence'' when models in the market are allowed to have some kind of interaction. Furthermore, though there are some theoretical efforts, it is still far from establishing a thorough theoretical framework for the learnware paradigm.

It is worth emphasizing that the learnware market is a fundamental infrastructure. Though it is possible to be built via volunteering service, in the long run enterprise-level learnware markets may be more favorable. This is because for learnware markets accommodating millions or even more learnwares, many involved issues such as the compressed storage, concurrency control, high throughput and low latency, need to be considered and the maintenance cost cannot be ignored. It would be beneficial to design credit incentive and versioning control mechanisms to encourage developers to submit upgraded models, and the learnware market can keep multiple versions of learnwares to provide cost-sensitive help to users; for this purpose, meta-data about versions and prices need to be maintained, in addition to the learnware models and specifications stored together in the market. It is expected that some learnware market enterprises may emerge. They build and maintain large-scale learnware markets, try to attract expert developers to submit excellent learners by giving credits such as monetary rewards, and get payment from end-users who get valuable help from the learnware market (the payment must be much less expensive than building a model from scratch by users themselves); this may give born to a learnware industry.

\section*{Acknowledgements}

This work was supported by the National Science Foundation of China (Grant No. 62250069).

\bibliographystyle{plain}
\bibliography{learnware}

\end{document}